\documentclass{article}
\usepackage{ijcai23}

\usepackage{times}
\usepackage{soul}
\usepackage{url}
\usepackage[hidelinks]{hyperref}
\usepackage[utf8]{inputenc}
\usepackage[small]{caption}
\usepackage{graphicx}
\usepackage{amsmath}
\usepackage{amsthm}
\usepackage{booktabs}
\usepackage{algorithm}
\usepackage{algorithmic}
\usepackage[switch]{lineno}

\usepackage{multirow}
\usepackage{amsfonts}       % blackboard math symbols
\usepackage{xcolor}         % colors
\usepackage{dsfont}
\usepackage{textcomp}
\usepackage{gensymb}
\usepackage{makecell}
\usepackage{enumitem}

\title{APR: Online Distant Point Cloud Registration through Aggregated Point Cloud Reconstruction}

\author{
Quan Liu$^1$
\and
Yunsong Zhou$^1$\and
Hongzi Zhu$^{1,}$\thanks{Corresponding author}\and
Shan Chang$^2$\and
Minyi Guo$^1$
\affiliations
$^1$Shanghai Jiao Tong University\\
$^2$Donghua University
\emails
\{liuquan2017,zhouyunsong,hongzi,guo-my\}@sjtu.edu.cn,
changshan@dhu.edu.cn
}
\begin{document}

\maketitle

\begin{abstract}
For many driving safety applications, it is of great importance to accurately register LiDAR point clouds generated on distant moving vehicles. However, such point clouds have extremely different point density and sensor perspective on the same object, making registration on such point clouds very hard.
In this paper, we propose a novel feature extraction framework, called \emph{APR}, for online distant point cloud registration.
Specifically, APR leverages an autoencoder design, where the autoencoder reconstructs a denser aggregated point cloud with several frames instead of the original single input point cloud. Our design forces the encoder to extract features with rich local geometry information based on one single input point cloud. Such features are then used for online distant point cloud registration.
We conduct extensive experiments against state-of-the-art (SOTA) feature extractors on KITTI and nuScenes datasets. Results show that APR outperforms all other extractors by a large margin, increasing average registration recall of SOTA extractors by 7.1\% on LoKITTI and 4.6\% on LoNuScenes. Code is available at \href{https://github.com/liuQuan98/APR}{https://github.com/liuQuan98/APR}.
\end{abstract}

\section{Introduction}
\label{sec:intro}
\begin{figure}[t]
  \centering
  \includegraphics[width=\linewidth]{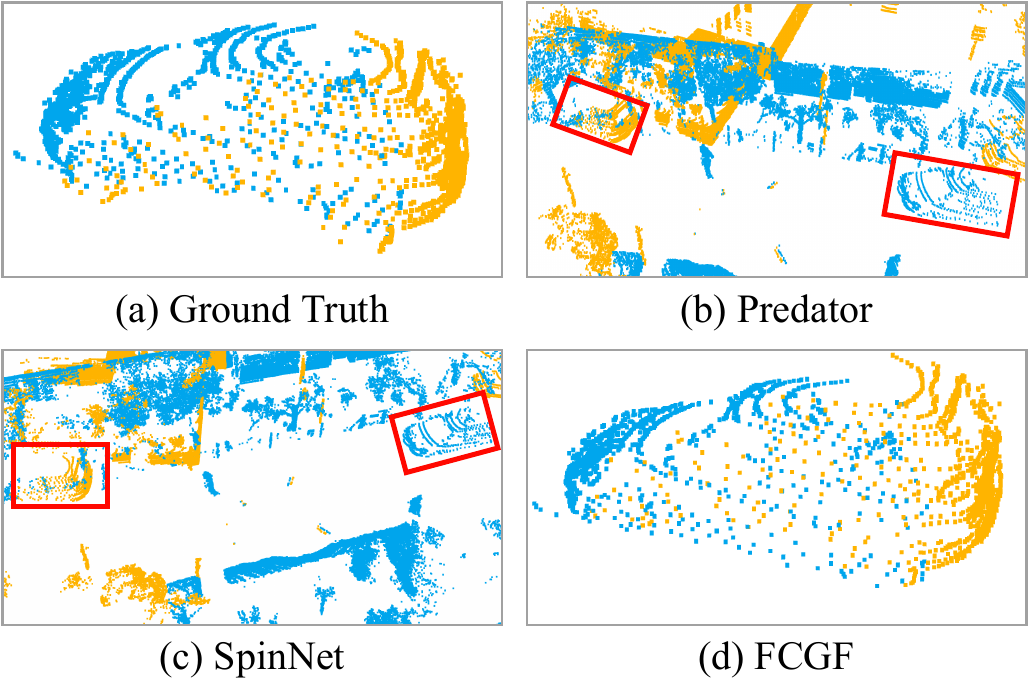}
  \caption{(a) An example vehicle extracted from two well-aligned point clouds with different point density and sensor perspective. (b)-(d) Registration results of SOTA point cloud registration methods, \emph{i.e.}, Predator, SpinNet and FCGF, respectively. Predator and SpinNet cannot align both point cloud well, with the example vehicle separated at two distinct locations indicated by the red boxes. FCGF can roughly align both point clouds but is still not accurate enough.}
  \label{fig:example}
\end{figure}

As LiDAR sensors have a precise and accurate 360\degree\ view, they are installed on new vehicle models for obstacle detection and avoidance to navigate safely. It is of great interest to share and align outdoor point clouds among neighboring vehicles via broadband wireless communication, which can greatly extend the visual field and improve the point density on distinct objects.
Because vehicles may be distant (\emph{e.g.} 20 to 50 meters apart), the corresponding point clouds are rather different in terms of point density and point of view about the same object in the scene. For example, Figure \ref{fig:example} (a) illustrates points about a target vehicle extracted from two nicely aligned point clouds obtained from two vehicles only 20 meters apart\footnote{Two point clouds separated by 20.48 meters are picked from the KITTI dataset for illustration without loss of generality.}, respectively. It can be seen that they have quite distinct perspectives about the target (\emph{i.e.} blue and orange points are obtained from the side front and the rear perspectives, respectively).
Despite such disparity of distant point clouds, if they can be well aligned, it would certainly enhance various downstream tasks such as object detection and semantic segmentation.

One practical online distant point cloud registration scheme for moving vehicles has to meet the following three requirements: 1) it has to be able to deal with the density and view disparity of such point clouds; 2) it should be cost efficient with respect to both computation and storage overhead for online inference; 3) it has to achieve superior accuracy as the result is crucial to autonomous or assisted driving decision and driving safety applications. For instance, it is extremely critical for driving safety applications to adopt a rigid registration criterion, \emph{e.g.}, 0.5\degree\ of rotation error and 0.5 meters of translation error (corresponding to a maximum translation error of less than one meter at all locations within 50 meters).

In the literature, most point cloud registration schemes target at \textit{close} point clouds (\emph{e.g.}, 10 meters apart), which have similar point density and share overlapped sensor perspective on target objects. Among these methods, feature extractors  \cite{choy2019fully,bai2020d3feat,ao2021spinnet,huang2021predator,poiesi2021distinctive} aim at improving feature quality. Outlier rejection methods  \cite{bai2021pointdsc,pais20203dregnet,choy2020deep} try to identify false correspondences. End-to-end methods in turn create new registration pipelines. These methods have made great effort in merging of traditional methods  \cite{aoki2019pointnetlk,yew2020rpm}, feature matching procedure  \cite{lu1905deepicp,li2020iterative,sarode2019pcrnet},  using transformer to extract contextual information  \cite{wang2019deep,wang2019prnet,ali2021rpsrnet}, and performing staged registration based on local patch similarity assumption \cite{lu2021hregnet,yu2021cofinet,qin2022geometric}. The accuracy when directly applying these methods to large-scale outdoor distant point clouds, however, is unsatisfactory. For example, Figure \ref{fig:example} (b)-(d) depict the results of SOTA point cloud registration methods on the example two distant point clouds shown in Figure \ref{fig:example} (a). As a result, to the best of our knowledge, there is no existing scheme successfully addressing the online distant point cloud registration problem.

In this paper, we propose a novel feature extraction framework, called \emph{Aggregated Point Cloud Reconstruction (APR)}, for online distant point cloud registration on moving vehicles. To defeat the density and view disparity of distant outdoor point clouds, it would be ideal if both vehicles have a full view and a better understanding about the same environment. 
Inspired by this insight, the main idea of APR is to train a powerful feature extractor by embedding a representation of a denser aggregated point cloud with a complete view about the current environment into the features, referred to as \emph{APR features}. Then, such features of two corresponding distant point clouds can be used for registration.

The APR design faces two main challenges as follows. First, training such a powerful feature extractor to contain a full view of the environment is non-trivial. In APR design, the extractor is trained via an autoencoder structure. Specifically, the encoder can be a state-of-the-art feature extraction backbone (\emph{e.g.}, FCGF or Predator), and metric learning loss is applied to the extracted features, making APR features preferable for registration based on feature similarity.
Furthermore, the decoder decodes the feature map of a single point cloud and derives a reconstructed point cloud, which is compared against the \emph{aggregated point cloud} (APC), defined as a series of point cloud frames of the corresponding vehicle aligned together. As a result, the capability of the encoder to guess a denser geometry is enhanced, so that APR feature contains rich environment information. Particularly, for memory-heavy backbones such as Predator, the encoder and decoder are asymmetrical to avoid out-of-memory (OOM) issue during training.

Second, it is challenging to perform online distant point cloud registration on two moving vehicles as it requires the registration algorithm to be not only accurate but fast.
Given a consecutive series of point cloud frames collected on each vehicle, a straightforward method is to take multiple frames from each vehicle as input to perform registration, which would incur heavy memory and computation costs since SLAM or multi-way registration is needed to register consecutive sweeps. In contrast, in our design, as the encoder has the capability to guess the features of APC, only two point cloud frames of a pair of vehicles are used to perform online pairwise registration. Point cloud series of each vehicle are only used to generate aggregated point clouds during offline training.

We implement APR on previous SOTA feature extractors, \emph{i.e.}, FCGF  \cite{choy2019fully} and Predator  \cite{huang2021predator}. We distill two low-overlap point cloud datasets, \emph{i.e.}, LoKITTI and LoNuScenes, with $\leq30\%$ overlap from KITTI and nuScenes and conduct extensive experiments. Results demonstrate that APR can effectively improve the performance of previous feature extractors, granting an increase of registration recall (RR) by 2.6\% on LoKITTI and 5.2\% on LoNuScenes when registering two point cloud frames ranging from 5 meters to 50 meters apart.

We highlight our main contributions made in this paper as follows:

\begin{itemize}[leftmargin=*]
    \item We propose an autoencoder design as a feature extraction framework, where the descriptiveness of the encoder could be enhanced despite the variety of decoder design, which is either symmetrical or asymmetrical.
    \item We introduce a new type of point cloud reconstruction target, where instead of using the input frame, we use several frames in vicinity to describe the environment in different views and densities, effectively confronting density variation and view disparity.
    \item We conduct extensive experiments and results demonstrate that APR achieves SOTA performance for distant point cloud registration.
\end{itemize}

%-------------------------------------------------------------------------

\section{Related Work}
\label{sec:relatedWork}

In this section, we first discuss traditional and learning-based feature extractors, which are closely related to our feature extractors; Then we move on to end-to-end methods and reconstruction-based methods to introduce recent progress on registration pipelines.

\subsection{Traditional Feature Extractors}
Traditional methods \cite{johnson1999using,rusu2009fast,tombari2010unique} represent earlier exploration of feature matching based on local shapes. SpinImages (SIs) \cite{johnson1999using} matches point clouds based on projection images; FPFH \cite{rusu2009fast} extracts rotation-invariant histograms of local geometries; SHOT \cite{tombari2010unique} functions by combining loca reference frames (LRFs) with geometry histograms. However, traditional methods usually ask for surface normal, which is hard to obtain in real-time scenarios, and are easily outperformed by learning methods due to their limited discriminative power.

\subsection{Learning-based Feature Extractors}

\paragraph{Patch-based Learning Methods.} The pioneering work for patch-based learning methods is 3DMatch \cite{zeng20173dmatch}, which applies 3D convolutions on local areas to extract local features for registration. PPF-Net \cite{deng2018ppfnet} utilizes PointNet \cite{qi2017pointnet} to extract robust point-pair features. PerfectMatch \cite{gojcic2019perfect} uses smoothed density value (SDV) to further improve feature robustness. Recent progress include DIP \cite{poiesi2021distinctive}, which uses a combinational loss of chamfer loss and hardest-contrastive loss \cite{choy2019fully}; SpinNet \cite{ao2021spinnet} combines LRF with SO(2) invariant convolution to achieve better rotation invariance. However, these methods generally have limited receptive field due to the use of local patches, and high computation time due to repeated local patch computation. They generally don't satisfy the real-time requirement for our problem.

\paragraph{Fully Convolutional Methods.} In order to naturally incorporate global and local information, fully convolutional methods are proposed. FCGF \cite{choy2019fully} first applies metric learning to dense convolutional features, achieving SOTA performance while being orders of magnitude faster than patch-based methods. D3Feat \cite{bai2020d3feat} improves upon FCGF through adopting KPConv \cite{thomas2019kpconv} and training a joint extractor-detector backbone. Predator \cite{huang2021predator} further approaches the low-overlap problem using overlap attention module at bottleneck, achieving SOTA performance. We implement our method based on these methods, as their fast and robust nature perfectly matches our requirements for a backbone.

\subsection{End-to-end Registration}
End-to-end registration methods generally modify components of the traditional registration pipeline, converting the pipeline into a network that can be trained in an end-to-end manner.

\paragraph{Superpoint Matching Methods.} Staged registration based on superpoints have seen success on high-overlap point clouds \cite{lu2021hregnet,yu2021cofinet,qin2022geometric}, where registration is first performed on downsampled point cloud (\emph{i.e.}, superpoints), then cascaded onto denser point cloud patches around the matched superpoints. However, these methods generally struggle to register distant point clouds, since the patch similarity assumption is broken under huge density variance and view disparity.

\paragraph{Other Methods.} Some methods such as PointNetLK \cite{aoki2019pointnetlk} and RPM-Net \cite{yew2020rpm} enhance optimization algorithms \cite{lucas1981iterative,gold1998new} with features extracted from PointNet \cite{qi2017pointnet}. Some methods such as DeepICP \cite{lu1905deepicp} and IDAM \cite{li2020iterative} refine the feature matching process in a learnable manner, reducing false positives. Other pipelines, including DCP \cite{wang2019deep}, PRNet \cite{wang2019prnet} and RPSRNet \cite{ali2021rpsrnet}, use transformers to grab the contextual information between features. However, they are generally dedicated to indoor or synthetic point cloud datasets, and are generally inable to scale up to the level of outdoor LiDAR point clouds.

\subsection{Reconstruction-based Registration}
We are aware of other methods that are also trying to merge chamfer loss and reconstruction into registration pipelines. Feature-metric Registration \cite{huang2020feature} uses the reconstruction of the input frame, which basically asks for no information loss. DIP \cite{poiesi2021distinctive} uses chamfer loss only for aligning two local patches. Compared to their focus on features rather than geometries, our decoder network directly learns the offset bias of the new points from features, focusing more on reconstructing local geometry.

%-------------------------------------------------------------------------

\section{Problem Modeling}
\label{sysmodel}

\subsection{System Model}
We consider conducting online distant point cloud registration on moving vehicles. Each vehicle is equipped with a LiDAR sensor and can continuously generate a time series of point cloud frames (\emph{e.g.}, ten point cloud frames per second) while moving. In addition, neighboring vehicles can efficiently exchange point cloud data in real time via broadband wireless communication techniques  \cite{wang2020demystifying,perfecto2017milimeter}. Vehicles have sufficient amount of memory for point cloud data storage but a restricted computational capability with an onboard embedded system. During offline training of our scheme, the rough distance estimation between a pair of point cloud frames in the point cloud series is required. It should be noted that, during online registration, we do not require any side channel information about how vehicles move via GPS or inertial sensors (\emph{e.g.}, accelerometers and gyroscopes).

\begin{figure*}[t]
    \centering
    \includegraphics[width=1\linewidth]{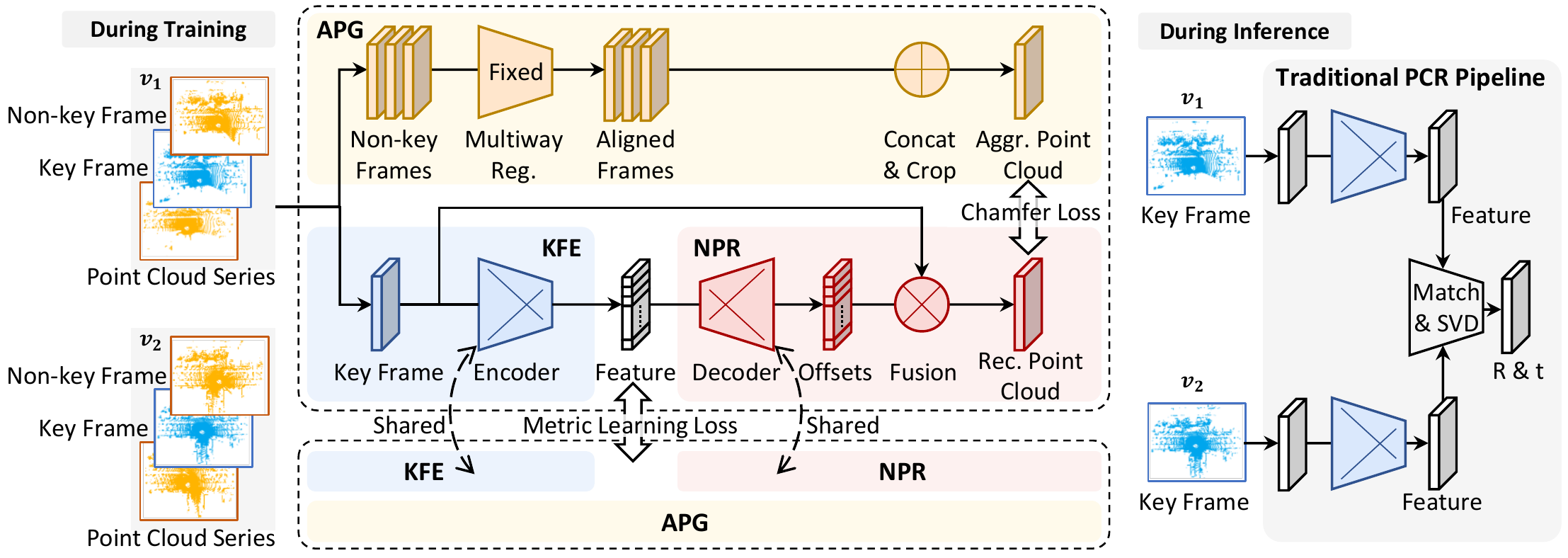}
    % \vspace{-0.6cm}
    \caption{The training and inference pipelines of APR. The offline training pipeline consists of three components, \emph{i.e.}, APG, KFE and NPR. APG generates the current aggregated point cloud (APC) based on nearby non-key frames of a vehicle in a non-trainable manner. KFE and NPR constitute an autoencoder, where the encoder extracts per-point features and the decoder takes such features to reconstruct the current APC. The Chamfer Distance between these two point clouds is used as a complementary loss against the metric learning loss to train the extractor (\emph{i.e.}, FCGF and Predator). During the online inference, only the current key frames of a pair of vehicles are exchanged and used to perform feature-based registration with the encoder at each vehicle.}
    \label{fig:architecture}
    % \vspace{-0.2cm}
\end{figure*}

\subsection{Problem Definition}
We denote the sequence $\{X_1^{v_i}, X_2^{v_i}, \dots, X_t^{v_i}\}$ as the time series of $t$ point cloud frames of vehicle $v_i$, where $X_k^{v_i}=\{p_i \in \mathbb{R}^3 | i=1,2,...,N\}$ for $k\in[1,t]$ is the $k$-th point cloud frame of $N$ points.
For a pair of vehicles $v_1$ and $v_2$ to register their current point cloud frame, \emph{e.g.}, $X_k^{v_1}=\{p_i \in \mathbb{R}^3 | i=1,2,...,N\}$ of $N$ points and $X_k^{v_2}=\{q_i \in \mathbb{R}^3 | i=1,2,...,M\}$ of $M$ points, the distant point cloud registration problem is to find the optimal rotation matrix $R\in SO(3)$ and translation vector $t\in \mathbb{R}^{3}$ that align the pair of point clouds, \emph{i.e.}, $X_k^{v_1}R^T+t^T$ aligns with $X_k^{v_2}$. For the ease of explanation, we refer to the point cloud frames to be registered as the \emph{key frames} and point cloud frames before and after one key frame as \emph{non-key frames}.

\section{System Design}
\label{systemdesign}

% \subsection{Overview}
The core idea of Aggregated Point Cloud Reconstruction (APR) is to leverage an autoencoder structure to train a powerful encoder as the feature extractor. Instead of reconstructing the original key frame, the decoder in the autoencoder structure reconstructs an aggregated point cloud. Therefore, with the well-trained encoder, features can be extracted from one single key frame but can effectively represent an environment point cloud with a dense and panoramic view, which are used for online registration. To this end, as depicted in Figure \ref{fig:architecture}, APR has distinct pipelines for offline training and for online inference, respectively.
More specifically, the training pipeline consists of the following three components:

\paragraph{Key-frame Feature Extraction (KFE).} KFE is the encoder in the autoencoder structure which is a fully convolutional network (FCN). The encoder acts as the feature extractor that extracts features from a key frame for both online registration during inference and non-key-frame reconstruction during training. APR is a universal feature extraction framework and can take an existing per-point feature extractor as its encoder. For instance, in our current implementation, two SOTA fully convolutional methods, \emph{i.e.}, FCGF  \cite{choy2019fully} and Predator  \cite{huang2021predator} are adopted to satisfy the real-time requirement for online registration. More point cloud feature extractors such as D3Feat \cite{bai2020d3feat} would be adopted in the future.

\paragraph{Aggregated Point Cloud Generation (APG).} APG takes neighboring non-key frames of the current key frame as input and aligns them to form an Aggregated Point Cloud (APC) as the reconstruction target of the decoder.

\paragraph{Non-key-frame Point Cloud Reconstruction (NPR).} NPR acts as the decoder of the autoencoder which explicitly extracts local geometries from the feature map by generating several new points around the point corresponding to a feature vector. The generated points are compared with the APC generated by the APG using the Chamfer Distance loss.

During online inference, only the current key frames are exchanged between a pair of vehicles. Then the key frames are used to extract APR features and perform conventional feature-based registration at each vehicle.

\subsection{Aggregated Point Cloud Generation}
\label{subsec:APG}
\begin{figure}[t]
    \centering
    % \vspace{-0.6cm}
    \includegraphics[width=0.8\linewidth]{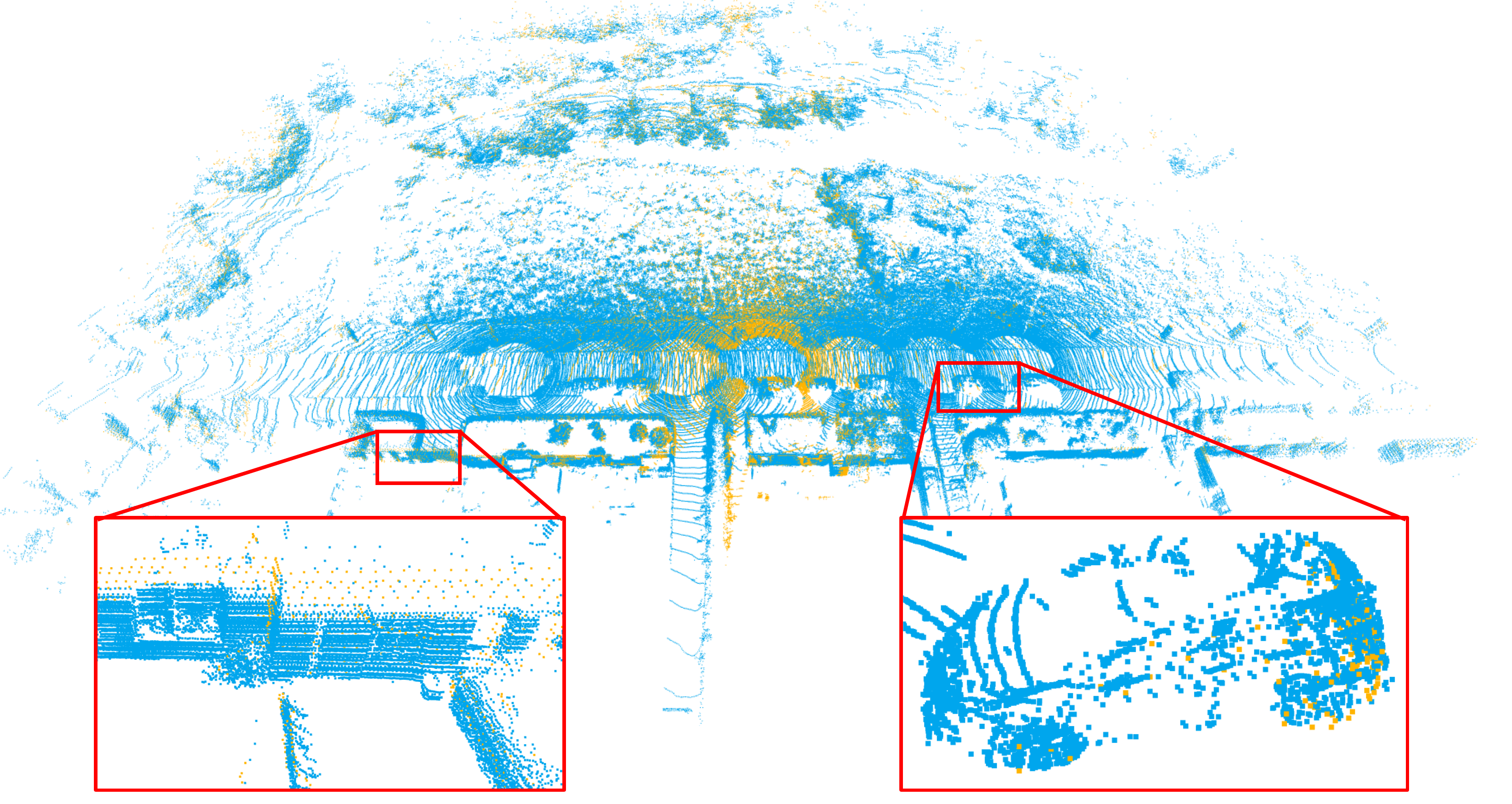}
    \caption{An example APC that consists of six non-key frames (points in blue), excluding the key frame (points in orange), with frames separated at a distance interval of ten meters. Objects are viewed from different perspectives in the generated APC, providing comprehensive shape information. In contrast, orange points of the key frame are more sparse and suffer from self-occlusion.}
    \label{fig:neighbourhoood}
    % \vspace{-0.2cm}
\end{figure}

The APG component is used to generate APC as the reconstruction targets for the autoencoder during offline training. Specifically, given the point cloud series of vehicle $v_i$, \emph{i.e.}, $\{X_1^{v_i}, X_2^{v_i}, \dots, X_t^{v_i}\}$ and a key frame $X_k^{v_i}$, APG first samples $2\psi$ non-key frames centered at the key frame $X_k^{v_i}$ from the point cloud series, \emph{i.e.}, $\psi$ frames before $X_k^{v_i}$ and $\psi$ frames after $X_k^{v_i}$ with frames separated at a short distance interval of $\alpha$ meters. Then, these sampled non-key frame and $X_k^{v_i}$ are aligned by ground-truth position label, or through a conventional multi-way registration method  \cite{Choi_2015_CVPR} when precise relative position is not available. The frames are further cropped with a scope sphere centered at the key frame to discard distant points out of interest. Finally, voxel downsampling is conducted on the APC to enhance the robustness against density variation. Figure \ref{fig:neighbourhoood} illustrates an APC using six non-key frames (denoted as blue points) centered at one key frame (denoted as orange points) with frames separated at a distance interval of ten meters.

\subsection{Non-key-frame Point Cloud Reconstruction}

Our key insight is that APC naturally suffers less from density variation and self-occlusion, as depicted in Figure \ref{fig:neighbourhoood}. Through the reconstruction of local geometry, ground truth correspondence features are encouraged to be similar in order to reconstruct similar local shape of APCs. This can be interpreted as a new side-channel for feature similarity supervision which is less influenced by density and view disparity.

Specifically, the encoder (feature extractor) in APR is trained to guess the local geometry and encode such information into a per-point feature map; The decoder in APR is trained to take the feature map, and decode it into a denser point cloud. Formally, given a key frame of $N$ points $X_k=\{p_i\in\mathbb{R}^3|i=1,2,...,N\}$ and the corresponding $l$-dimensional features $F=\{f_i\in\mathbb{R}^l | i=1,2,...,N\}$, the decoder can be formulated in two ways, either symmetrical or asymmetrical.

\paragraph{Symmetrical Design.} In this case, the decoder $D_{sym}$ is an FCN that shares identical structure with the encoder except for the input and output dimensions. The decoder takes the feature map and outputs the reconstructed offsets as the output feature $O=D_{sym}(F)\in \mathbb{R}^{N\times 3\phi}$, where $\phi$ is referred to as the \emph{point generation ratio}.

\paragraph{Asymmetrical Design.} Alternatively, the decoder $D_{asym}$ can be a small MLP that takes a single feature vector as input, providing a lightweight and universal option. For each feature $f_i$, the decoder locally generates the set of location offsets as output $O_i=D_{asym}(f_i)\in\mathbb{R}^{3\phi}$.

\paragraph{Fusion.} We interpret $O_i$ as $\phi$ segments of 3-dimensional vectors with each vector being an offset $b_i^1, b_i^2, ..., b_i^\phi \in \mathbb{R}^3$. By adding these location offsets to the corresponding original point $p_i$ in the key frame, a reconstructed point cloud is obtained: $R_k = \{p_i + b_i^j \in \mathbb{R}^3 | i=1,2,...,N; j=1,2,...,\phi\}$. The collected offsets are denoted as $\hat{O}=\{b_i^j \in \mathbb{R}^3 | i=1,2,...,N; j=1,2,...,\phi\}$. We contract the index i and j into a single index in the following section for convenience of display.

\subsection{Loss Formulation}
To train the autoencoder, the Chamfer Distance (CD) between the APC generated by the APG and the reconstructed environment generated by the NPR is used as a complementary loss against the metric learning loss involving the extracted features from a pair of point clouds. L2 regularization is also applied to the generated offset lengths.

Specifically, given a pair of point clouds $X_k^{v_1}=\{p_i \in \mathbb{R}^3\}$, $X_k^{v_2}=\{q_j \in \mathbb{R}^3\}$, and their corresponding offsets $\hat{O}^{v_1}=\{o_l \in \mathbb{R}^3\}$, $\hat{O}^{v_2}=\{o_m \in \mathbb{R}^3\}$, CD loss and regularization loss are formulated as:
\begin{equation}
\begin{split}
    L_{CD} = \frac{1}{|X_k^{v_1}|} \sum\limits_{p_i\in X_k^{v_1}}{\min\limits_{q_j\in X_k^{v_2}}||p_i-q_j||^2}\\
           + \frac{1}{|X_k^{v_2}|} \sum\limits_{q_j\in X_k^{v_2}}{\min\limits_{p_i\in X_k^{v_1}}||p_i-q_j||^2}
\end{split}
\label{chamferdistance}
\end{equation}

\begin{equation}
\begin{split}
    L_{L2} = \frac{1}{|\hat{O}^{v_1}|} \sum\limits_{o_l\in \hat{O}^{v_1}}{||o_l||^2}
            + \frac{1}{|\hat{O}^{v_2}|} \sum\limits_{o_m\in \hat{O}^{v_2}}{||o_m||^2}
\end{split}
\label{l2regularization}
\end{equation}

Metric Learning (ML) is generally utilized to train effective fully-convolutional feature extractors. In this paper, we use the original Hardest Contrastive Loss for FCGF \cite{choy2019fully}, and the Circle Loss \cite{sun2020circle} plus Overlap Loss and Matchability Loss for Predator \cite{huang2021predator}, same as their original paper.

Let $L_{ML}$ denote the metric learning loss of an embedded point cloud feature extraction scheme (\emph{e.g.}, FCGF  \cite{choy2019fully} and Predator  \cite{huang2021predator}) and the final loss is formulated as $L = L_{ML} + \lambda_1 L_{CD} + \lambda_2 L_{L2}$, where $\lambda_1, \lambda_2$ controls the ratio between loss terms.

\section{Performance Evaluation}
\label{evaluation}

\paragraph{Distant Point Cloud Datasets.} Previously, only close-range registration datasets have been extracted from KITTI \cite{Geiger2012CVPR} and nuScenes \cite{Caesar_2020_CVPR}. To create distant point cloud datasets, following 3DLoMatch proposed in Predator \cite{huang2021predator}, we pick point cloud pairs with $\leq30\%$ overlap from KITTI and nuScenes, creating LoKITTI and LoNuScenes, respectively. We also subdivide KITTI and nuScenes \emph{w.r.t.} the distance between two LiDAR centers, denoted as $d$ (in meters). Dataset slices with $d\in[d_1, d_2]$ in KITTI and nuScenes are dubbed KITTI$[d_1, d_2]$ and nuScenes$[d_1,d_2]$, respectively. In addition, previous close-range datasets with $d=10$, referred to as KITTI and nuScenes, are also considered. We adopt three registration criteria for assessing a successful registration, \emph{i.e.}, \emph{loose} ($5^\circ$ \& 2m)  \cite{yew20183dfeat}, \emph{normal} ($1.5^\circ$ \& 0.6m) and \emph{strict} ($0.5^\circ$ \& 0.3m). We report three metrics, including relative rotation error (RRE), relative translation error (RTE), and registration recall (RR). For alignment of non-key frames in APG, we use only ground-truth pose from SemanticKITTI  \cite{behley2019iccv} and nuScenes  \cite{Caesar_2020_CVPR}, while the effect of low-precision position labels are simulated in the robustness analysis in Secion \ref{robustness_analysis}. We set default parameters as $\psi=3$ and $\alpha=10$.

\begin{table*}[ht]
  \centering
  \small
  \resizebox{0.85\linewidth}{4cm}{
  \begin{tabular}{@{}c|lccc|ccc|ccc@{}}
    \toprule
    \multicolumn{2}{c}{}& \multicolumn{3}{c|}{Loose Criterion (5\degree,2m)} & \multicolumn{3}{c|}{Normal Criterion (1.5\degree,0.6m)} & \multicolumn{3}{c}{Strict Criterion (0.5\degree,0.3m)} \\
    \midrule
    \multicolumn{1}{c}{}&Metrics &RRE (\degree) &RTE (cm) & RR (\%)  &RRE (\degree) &RTE (cm) & RR (\%) &RRE (\degree) &RTE (cm) & RR (\%)  \\
    \midrule
    \multirow{5}{*}{\rotatebox{90}{KITTI}} &FCGF            &0.35 	&10.6 	&98.2 	&\bf{0.27} 	&8.1 	&96.9 	&\bf{0.20} 	&7.8 	&85.3 \\
    &FCGF+APR(a)      &0.34 	&9.6 	&98.2 	&0.30 	&9.6 	&97.1 	&0.22 	&8.6 	&86.9 \\
    &FCGF+APR(s)      &\bf{0.30} 	&10.0 	&99.0 	&0.28 	&9.4 	&97.1 	&0.21 	&9.2 	&84.7 \\
    &Predator        &0.31 	&7.4 	&\bf{100.0} 	&0.30 	&7.4 	&98.5 	&0.24 	&\bf{7.2} 	&85.3 \\
    &Predator+APR(a)  &\bf{0.30} 	&\bf{7.3} 	&\bf{100.0} 	&0.28 	&\bf{7.3} 	&\bf{99.0} 	&0.23 	&\bf{7.2} 	&\bf{88.0} \\
    \midrule
    \multirow{5}{*}{\rotatebox{90}{LoKITTI}}&FCGF            &2.02 	&55.2 	&22.2 	&0.91 	&32.1 	&5.1 	&0.42 	&19.0 	&1.3 \\
    &FCGF+APR(a)      &1.74 	&51.9 	&32.7 	&\bf{0.74} 	&28.6 	&18.9 	&\bf{0.31} 	&\bf{18.4} 	&3.1 \\
    &FCGF+APR(s)      &1.76 	&47.9 	&33.0 	&0.90 	&\bf{27.5} 	&16.3 	&0.37 	&19.1 	&2.8 \\
    &Predator        &1.75 	&43.4 	&42.4 	&0.83 	&29.9 	&22.0 	&0.35 	&18.5 	&4.2 \\
    &Predator+APR(a)  &\bf{1.64} 	&\bf{39.5} 	&\bf{50.8} 	&0.83 	&27.8 	&\bf{29.9} 	&0.34 	&\bf{18.4} 	&\bf{5.5} \\
    \midrule
    \multirow{5}{*}{\rotatebox{90}{nuScenes}}&FCGF            &0.46 	&50.0 	&93.6 	&\bf{0.38} 	&20.0 	&77.9 	&0.27 	&14.8 	&55.9 \\
    &FCGF+APR(a)      &\bf{0.45} 	&37.0 	&94.5 	&0.39 	&18.2 	&78.1 	&0.27 	&14.1 	&55.1 \\
    &FCGF+APR(s)      &0.47 	&47.0 	&98.4 	&0.43 	&20.6 	&78.0 	&\bf{0.23} 	&14.7 	&53.2 \\
    &Predator        &0.58 	&20.2 	&97.8 	&0.52 	&18.0 	&92.3 	&0.32 	&14.0 	&44.5 \\
    &Predator+APR(a)  &0.47 	&\bf{19.1} 	&\bf{99.5} 	&0.45 	&\bf{17.1} 	&\bf{95.3} 	&0.31 	&\bf{13.8} 	&\bf{56.9} \\
    \midrule
    \multirow{5}{*}{\rotatebox{90}{LoNuScenes}} &FCGF            &\bf{1.30} 	&60.9 	&49.1 	&\bf{0.72} 	&30.0 	&23.7 	&0.34 	&18.5 	&7.8 \\
    &FCGF+APR(a)      &1.40 	&62.0 	&51.8 	&0.68	&29.9 	&23.6 	&\bf{0.33}	&\bf{18.3} 	&8.4 \\
    &FCGF+APR(s)      &1.35 	&65.8 	&50.8 	&0.72	&28.8 	&24.3 	&0.34	&18.9 	&8.2 \\
    &Predator        &1.47 	&54.5 	&50.4 	&0.78	&31.0 	&26.3 	&0.34	&18.9 	&5.3 \\
    &Predator+APR(a)  &\bf{1.30} 	&\bf{51.8} 	&\bf{62.7} 	&\bf{0.72}	&\bf{29.8} 	&\bf{35.3} 	&0.34	&19.2 	&\bf{8.5} \\
    \bottomrule
  \end{tabular}
  }
%   \vspace{-0.1cm}
  \caption{Comparison of FCGF and Predator with their APR-empowered version on KITTI, LoKITTI, nuScenes and LoNuScenes under different registration criteria. Symmetrical and asymmetrical designs are tagged as '(a)' and '(s)', respectively.}
  \label{tab:ovarallcomparison}
%   \vspace{-0.4cm}
\end{table*}

\paragraph{Training Strategy.} Although we report arbitrary results on $[d_1,d_2]$ datasets, it is hard for the network to converge when directly trained with $d_2\geq30$ or $d_1\geq10$. As a result, we first pre-train a model on a dataset with lower distance, where $d\in[5, 20]$. Then the pre-trained model is further finetuned on $[5,d_2]$ ($d_2\geq30$) to guarantee convergence. For cases where $d_2<30$, no finetuning is applied. Finally, the resulting model is used to report results on $[d_1,d_2]$ datasets.

\paragraph{Decoder Choice.} We implement both symmetrical design (denoted with postfix $(s)$) and asymmetrical design (denoted with postfix $(a)$) on FCGF, and found that symmetrical design generally performs better. However, symmetrical Predator is not trainable due to memory constraints, so only asymmetrical design is reported for Predator.

\begin{figure*}[t]
    \centering
    \includegraphics[width=1\linewidth]{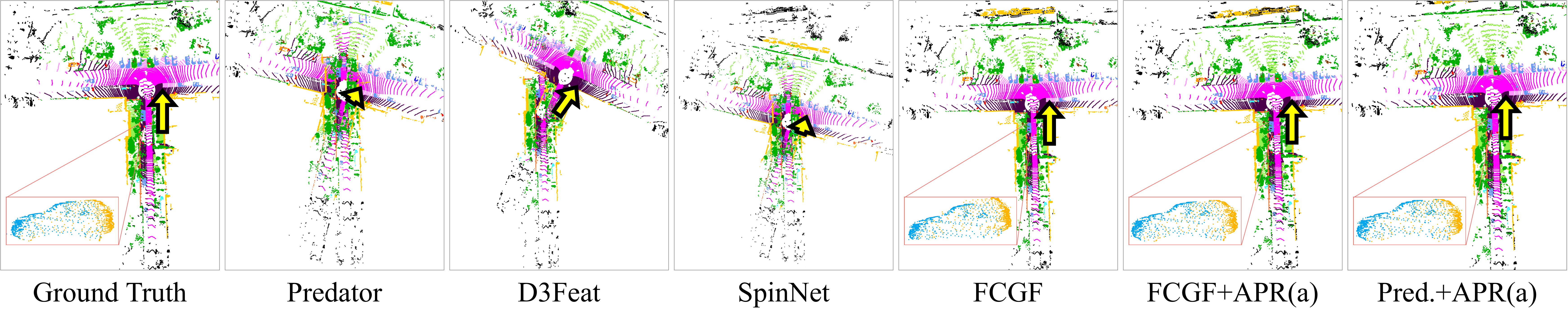}
    % \vspace{-0.6cm}
    \caption{Qualitative distant point cloud registration results on KITTI$[5,20]$. The yellow arrow indicates the relative transformation between this pair of point clouds. Both APR-empowered methods perform perfectly while FCGF actually fails the loose registration criterion with 5.32\degree\ rotation error, as highlighted with the displacement of the example vehicle.}
    \label{fig:qualitative}
    % \vspace{-0.3cm}
\end{figure*}

\paragraph{Visualization.} We demonstrate the registration performance of all methods on KITTI in Figure \ref{fig:qualitative}.

\subsection{Parameter Configuration}
\label{parameterConfig}

\paragraph{Effect of Point Generation Ratio $\phi$.} We first examine the effect of the point generation ratio $\phi$, varying it from one to eight. Table \ref{tab:pgr_decoder} lists the performance of FCGF+APR(a/s) and Predator+APR(a) on KITTI$[5, 20]$ validation set. It can be seen that the performance of FCGF+APR(a/s) peaks when $\phi=4$, while the performance of Predator+APR drops when $\phi<4$. We can see that $\phi=4$ is a good choice for all feature extractors.

\paragraph{Effect of Decoder Size.} We fix the point generation ratio $\phi$ to four and vary the size of the per-point decoder in asymmetrical designs. Specifically, the asymmetrical decoder is an MLP with flexible hidden layer size, \emph{e.g.}, $(2^9, 2^8)$ reveals a 3-layer MLP with $l$, 512, 256, $\phi\times3$ dimensions from input to output. Table \ref{tab:pgr_decoder} lists the performance of FCGF+APR(a) and Predator+APR(a) on KITTI$[5, 20]$ validation set. It can be seen that the decoder with the size of $(2^9, 2^8)$ can achieve the best RR performance. Therefore, we set the decoder hidden layer size in asymmetrical design to $(2^9, 2^8)$ in the following experiments.

% \paragraph{Other Parameters} We refer all readers to the Appendix for the choice of non-key frame number $\psi$ and all parameter configurations on nuScenes.

\begin{table}[t]
  \centering
  \small
  \resizebox{1\linewidth}{2.3cm}{
  \begin{tabular}{@{}lccc|ccc|ccc@{}}
    \toprule
     & \multicolumn{3}{c|}{FCGF+APR(a)} & \multicolumn{3}{c|}{FCGF+APR(s)} & \multicolumn{3}{c}{Predator+APR(a)} \\
    \midrule
    PGR$(\phi)$  &RRE &RTE & RR &RRE &RTE & RR &RRE &RTE & RR \\
    \midrule
    1 & 0.32 & 9.3 & 92.4 & 0.33 & 10.1 & 94.4 & 0.41 &9.8 &98\\
    2 & 0.31 & 8.8 & 93.0 & 0.34 & 12.0 & 94.6 & 0.42 &9.6 &99\\
    4 & 0.31 & 8.7 & 93.0 & 0.31 & 10.7 & 95.1 &0.37  &8.7  &100\\
    8 & 0.30 & 10.0 & 92.8 & 0.32 & 10.6 & 93.2 &0.34 &8.8 &100\\
    \midrule\midrule
    Decoder size  & RRE &RTE & RR &RRE &RTE & RR &RRE &RTE & RR \\
    \midrule
    $(2^4)$         &0.33	&11.8	&96.1 &\multicolumn{3}{c|}{\multirow{4}{*}{None (symmetrical)}}&0.31	&8.1	&97.2\\
    $(2^5, 2^4)$    &0.32	&12.1	&96.4 &&&&0.31	&8.2	&96.8\\
    $(2^9, 2^8)$    &0.33 	&11.9 	&96.4 &&&&0.32 	&8.1    &97.3\\
    $(2^{11}, 2^{10}, 2^9)$ &0.35 	&9.1&96.1 &&&&0.31	&8.2   &97.3\\
    \bottomrule
  \end{tabular}
  }
%   \vspace{-0.2cm}
  \caption{Performance of FCGF+APR(a/s) and Predator+APR(a) with different point generation ratio $\phi$ (upper half) and decoder hidden layer size (lower half) on KITTI$[5, 20]$ dataset under normal registration criterion. Decoder size of FCGF+APR(s) remain the same as the encoder and stay unchanged.}
  \label{tab:pgr_decoder}
%   \vspace{-0.3cm}
\end{table}

\subsection{Performance Comparison}
\label{comparison}

\paragraph{Improvements over Baselines.} We compare our method with baseline feature extractors under all three criteria on all four datasets. Table \ref{tab:ovarallcomparison} lists the registration results. Compared to the baselines, APR can effectively improve the registration performance both under normal overlap and low overlap, either symmetrical or asymmetrical. Symmetrical designs excel under loose criterion but lose advantage under rigid criterion. Though Predator+APR(a) does not excel in all tests with RRE or RTE slightly falling behind in some cases, it generally outperforms other methods in most scenarios.
On average, APR improves the RR of FCGF and Predator by 4.4\%, 3.5\%, respectively on KITTI, and 0.6\%, 6.9\%, respectively on nuScenes.
% Predator+APR(a) can achieve a maximum of 12.4\% RR gain over Predator on LoNuScenes dataset under the loose registration criterion.
The average RR improvements for both backbones on LoKITTI and LoNuScenes are 7.1\% and 4.6\%, respectively.

\paragraph{Impact of Distance $d$ for all Methods.} Due to the nature of LiDAR sensors, points on near objects are denser than those on distant objects. Consequently, density and view disparity are mainly caused by increased distance between both LiDARs. Table \ref{tab:distance} shows the RR of all candidate methods under increasing ranges of $d$ on KITTI$[d_1, d_2]$ dataset under normal registration criterion  (1.5\degree,0.6m). It can be seen that SpinNet, D3Feat and Geotransformer gradually fail with increasing $d$, as they suffer from bad convergence caused by improper loss or structure designs. The APR-empowered methods generally receive a performance boost that is proportional with the distance $d$, and Predator+APR(s) surpasses all other methods on all distances. We conclude that APR can effectively confront the density and view disparity between two separated point clouds, and sets a new SOTA for distant point cloud registration problem. 
% Due to page limitations, other metrics of all methods under all distance $d$ are placed in the Appendix.

\begin{table}[ht]
  \centering
  \small
  \resizebox{1\linewidth}{2cm}{
  \begin{tabular}{@{}lccccc@{}}
    \toprule
     Range of $d$ (m) & $[5, 10]$ & $[10, 20]$ & $[20, 30]$ & $[30, 40]$ & $[40, 50]$  \\
    \midrule
    SpinNet	&97.6	&73.1	&7.3	&0	&0\\
    D3Feat	&98.7	&86.8	&52.7	&20	&4.5\\
    CoFiNet	&\bf{99.6} 	&94.2 	&80.0 	&44.8 	&24.3 \\
    GeoTransformer	&97.9 	&88.3 	&8.3 	&0.7 	&0.0\\ 
    FCGF	&97.0 	&85.4 	&54.1 	&25.0 	&14.3\\ 
    FCGF+APR(a)	&97.5 	&88.9 	&55.7 	&35.0 	&17.1 \\
    FCGF+APR(s)	&98.4 	&93.0 	&58.4	&42.1	&22.5\\
    Predator	&99.3	&96.8	&90.2	&60.6	&26.7\\
    Predator+APR(a)	&\bf{99.6} 	&\bf{97.4} 	&\bf{91.2} 	&\bf{62.1} 	&\bf{40.8} \\
    \bottomrule
  \end{tabular}
  }
  \caption{RR ($\%$) of all methods on 5 KITTI$[d_1, d_2]$ datasets with different ranges of $d\in[d_1, d_2]$ under normal registration criterion.}
  \label{tab:distance}
\end{table}

\label{robustness_analysis}
\paragraph{Robustness Analysis.} We investigate the robustness of APR, assuming APG could fail to align some non-key frames with only noisy pose annotation available, as in OdometryKITTI. To simulate different degrees of failure during training, we impose random rotation ($\theta\in[-\pi,\pi]$) on a random axis on $N_{disturb}$ out of $2\phi$ non-key frames. Table \ref{tab:mutateNeighbour} lists the RR of APR on both KITTI$[5, 20]$ and nuScenes$[5, 20]$ dataset with $N_{disturb}\in[0,5]$. Under the optimal parameter setting, we have $\psi=3$, \emph{i.e.}, 6 non-key frames, 3 on each side. It can be seen that asymmetrical design performs better, whose RR first rises then decreases with $N_{disturb}$, peaking at $95.4\%$ for FCGF+APR(a), and $99.5\%$ for Predator+APR(a), respectively on KITTI dataset. Similar trends can be observed on nuScenes dataset. When $N_{disturb}\geq4$, RR for both asymmetrical methods drops slightly. All versions of APR, are comparable with not-disturbed version ($N_{disturb}=0$), and better than native methods (see the last column). We conclude that APR is robust against APG failures.

\begin{table}[t]
  \centering
  \small
  \resizebox{\linewidth}{2cm}{
  \begin{tabular}{@{}lcccccc|c@{}}
    \toprule
    $N_{disturb}$  & 0 & 1 & 2 & 3 & 4 & 5 & /\\
    \midrule
    &\multicolumn{6}{c}{KITTI$[5, 20]$} \\
    \midrule
    FCGF+APR(a)      &94.3	&94.2	&94.2	&95.4	&93.7	&93.4   &87.6\\
    FCGF+APR(s)      &95.1  &87.8	&93.2	&91.3	&93.2	&93.2   &87.6\\
    Predator+APR(a)  &99.0   &98.5	&99.5	&98.3	&98.1	&98.5   &98.8\\
    \midrule
    &\multicolumn{6}{c}{nuScenes$[5, 20]$} \\
    \midrule
    FCGF+APR(a)      &61.2	&58.1	&61.2	&62.1	&63.3	&61.1	&53.6\\
    FCGF+APR(s)      &54.0	&55.7	&54.4	&53.7	&55.9	&56.7	&53.6\\
    Predator+APR(a)  &85.2	&82.3	&83.4	&86.1	&72.1	&70.8	&65.9\\
    \bottomrule
  \end{tabular}
  }
  \caption{RR for APR-empowered methods under normal registration criterion, with $d\in[5, 20]$ and 6 non-key frames for each key frame. $N_{disturb}$ out of 6 non-key frames are randomly rotated, providing noise to APR instead of useful information. The '/' column shows results of raw FCGF and Predator.}
  \label{tab:mutateNeighbour}
\end{table}

\begin{table}[t]
  \centering
  \small
  \resizebox{0.7\linewidth}{1.5cm}{
  \begin{tabular}{@{}lcccc@{}}
    \toprule
    Downsample Ratio & 0.1 & 0.2 & 0.5 & 1 \\
    \midrule
    % &\multicolumn{4}{c}{nuScenes$[5, 10]$} \\
    % \midrule
    FCGF         &44.6   &57.6   &70.4   &78.4\\
    FCGF+APR(a)      &49.4	&63.1	&76.2	&85.3\\
    Relative $\Delta$RR (\%)        &10.7  &9.55   &8.81   &8.80\\
    \midrule
    Predator       &3.4	&32	    &80.2	&96.6\\
    Predator+APR(a)  &3.8	&34.4	&84.9	&98.1\\
    Relative $\Delta$RR (\%)        &11.7	&7.50	&5.86	&1.55\\
    \bottomrule
  \end{tabular}
  }
  \caption{RR for asymmetrical APR and the relative RR improvement compared with FCGF and Predator on nuScenes$[5,10]$ dataset, with one point cloud out of every pair downsampled by a certain ratio.}
  \label{tab:density}
\end{table}

\paragraph{Impact of Density Variation.} In this experiment, we examine the effect of density variation alone, without view disparity. We choose $d\in[5,10]$ to minimize the difference in point of view, then randomly downsample one point cloud out of every point cloud pair, creating simulated density variation between a pair of point clouds. Table \ref{tab:density} lists the RR of FCGF, Predator and their asymmetrical APR-empowered version on nuScenes$[5,10]$ dataset under normal registration criterion. The relative improvement of RR for APR-empowered methods compared to native methods increases as the downsample ratio decreases, meaning that APR is effective in confronting point cloud density variation.

\section{Conclusion}
\label{conclusion}
We have proposed APR, a novel feature extraction framework for online distant point cloud registration on moving vehicles. APR leverages an autoencoder design to obtain a better feature extractor through reconstruction of the aggregated point clouds during training. Our method is able to force the feature extractor to guess local geometry information without changing any of the extractor design or imposing extra inference time. The extracted APR features are more robust against density variation and view disparity, significantly improving the accuracy of distant point cloud registration. APR outperforms all other extractors by a large margin, increasing average registration recall of SOTA extractors by 7.1\% on LoKITTI and 4.6\% on LoNuScenes.

\bibliographystyle{named}
\bibliography{ijcai23}

\begin{thebibliography}{}

\bibitem[\protect\citeauthoryear{Ali \bgroup \em et al.\egroup
  }{2021}]{ali2021rpsrnet}
Sk~Aziz Ali, Kerem Kahraman, Gerd Reis, and Didier Stricker.
\newblock Rpsrnet: End-to-end trainable rigid point set registration network
  using barnes-hut 2d-tree representation.
\newblock In {\em Proceedings of the IEEE/CVF Conference on Computer Vision and
  Pattern Recognition}, pages 13100--13110, 2021.

\bibitem[\protect\citeauthoryear{Ao \bgroup \em et al.\egroup
  }{2021}]{ao2021spinnet}
Sheng Ao, Qingyong Hu, Bo~Yang, Andrew Markham, and Yulan Guo.
\newblock Spinnet: Learning a general surface descriptor for 3d point cloud
  registration.
\newblock In {\em Proceedings of the IEEE/CVF Conference on Computer Vision and
  Pattern Recognition}, pages 11753--11762, 2021.

\bibitem[\protect\citeauthoryear{Aoki \bgroup \em et al.\egroup
  }{2019}]{aoki2019pointnetlk}
Yasuhiro Aoki, Hunter Goforth, Rangaprasad~Arun Srivatsan, and Simon Lucey.
\newblock Pointnetlk: Robust \& efficient point cloud registration using
  pointnet.
\newblock In {\em Proceedings of the IEEE/CVF Conference on Computer Vision and
  Pattern Recognition}, pages 7163--7172, 2019.

\bibitem[\protect\citeauthoryear{Bai \bgroup \em et al.\egroup
  }{2020}]{bai2020d3feat}
Xuyang Bai, Zixin Luo, Lei Zhou, Hongbo Fu, Long Quan, and Chiew-Lan Tai.
\newblock D3feat: Joint learning of dense detection and description of 3d local
  features.
\newblock In {\em Proceedings of the IEEE/CVF Conference on Computer Vision and
  Pattern Recognition}, pages 6359--6367, 2020.

\bibitem[\protect\citeauthoryear{Bai \bgroup \em et al.\egroup
  }{2021}]{bai2021pointdsc}
Xuyang Bai, Zixin Luo, Lei Zhou, Hongkai Chen, Lei Li, Zeyu Hu, Hongbo Fu, and
  Chiew-Lan Tai.
\newblock Pointdsc: Robust point cloud registration using deep spatial
  consistency.
\newblock In {\em Proceedings of the IEEE/CVF Conference on Computer Vision and
  Pattern Recognition}, pages 15859--15869, 2021.

\bibitem[\protect\citeauthoryear{Caesar \bgroup \em et al.\egroup
  }{2020}]{Caesar_2020_CVPR}
Holger Caesar, Varun Bankiti, Alex~H. Lang, Sourabh Vora, Venice~Erin Liong,
  Qiang Xu, Anush Krishnan, Yu~Pan, Giancarlo Baldan, and Oscar Beijbom.
\newblock nuscenes: A multimodal dataset for autonomous driving.
\newblock In {\em Proceedings of the IEEE/CVF Conference on Computer Vision and
  Pattern Recognition}, June 2020.

\bibitem[\protect\citeauthoryear{Choi \bgroup \em et al.\egroup
  }{2015}]{Choi_2015_CVPR}
Sungjoon Choi, Qian-Yi Zhou, and Vladlen Koltun.
\newblock Robust reconstruction of indoor scenes.
\newblock In {\em Proceedings of the IEEE/CVF Conference on Computer Vision and
  Pattern Recognition}, June 2015.

\bibitem[\protect\citeauthoryear{Choy \bgroup \em et al.\egroup
  }{2019}]{choy2019fully}
Christopher Choy, Jaesik Park, and Vladlen Koltun.
\newblock Fully convolutional geometric features.
\newblock In {\em Proceedings of the IEEE/CVF International Conference on
  Computer Vision}, 2019.

\bibitem[\protect\citeauthoryear{Choy \bgroup \em et al.\egroup
  }{2020}]{choy2020deep}
Christopher Choy, Wei Dong, and Vladlen Koltun.
\newblock Deep global registration.
\newblock In {\em Proceedings of the IEEE/CVF Conference on Computer Vision and
  Pattern Recognition}, pages 2514--2523, 2020.

\bibitem[\protect\citeauthoryear{Deng \bgroup \em et al.\egroup
  }{2018}]{deng2018ppfnet}
Haowen Deng, Tolga Birdal, and Slobodan Ilic.
\newblock Ppfnet: Global context aware local features for robust 3d point
  matching.
\newblock In {\em Proceedings of the IEEE/CVF Conference on Computer Vision and
  Pattern Recognition}, pages 195--205, 2018.

\bibitem[\protect\citeauthoryear{Geiger \bgroup \em et al.\egroup
  }{2012}]{Geiger2012CVPR}
Andreas Geiger, Philip Lenz, and Raquel Urtasun.
\newblock Are we ready for autonomous driving? the kitti vision benchmark
  suite.
\newblock In {\em Proceedings of the IEEE/CVF Conference on Computer Vision and
  Pattern Recognition}, 2012.

\bibitem[\protect\citeauthoryear{Gojcic \bgroup \em et al.\egroup
  }{2019}]{gojcic2019perfect}
Zan Gojcic, Caifa Zhou, Jan~D Wegner, and Andreas Wieser.
\newblock The perfect match: 3d point cloud matching with smoothed densities.
\newblock In {\em Proceedings of the IEEE/CVF Conference on Computer Vision and
  Pattern Recognition}, pages 5545--5554, 2019.

\bibitem[\protect\citeauthoryear{Gold \bgroup \em et al.\egroup
  }{1998}]{gold1998new}
Steven Gold, Anand Rangarajan, Chien-Ping Lu, Suguna Pappu, and Eric Mjolsness.
\newblock New algorithms for 2d and 3d point matching: pose estimation and
  correspondence.
\newblock {\em Pattern Recognition}, 31(8):1019--1031, 1998.

\bibitem[\protect\citeauthoryear{Huang \bgroup \em et al.\egroup
  }{2020}]{huang2020feature}
Xiaoshui Huang, Guofeng Mei, and Jian Zhang.
\newblock Feature-metric registration: A fast semi-supervised approach for
  robust point cloud registration without correspondences.
\newblock In {\em Proceedings of the IEEE/CVF Conference on Computer Vision and
  Pattern Recognition}, pages 11366--11374, 2020.

\bibitem[\protect\citeauthoryear{Huang \bgroup \em et al.\egroup
  }{2021}]{huang2021predator}
Shengyu Huang, Zan Gojcic, Mikhail Usvyatsov, Andreas Wieser, and Konrad
  Schindler.
\newblock Predator: Registration of 3d point clouds with low overlap.
\newblock In {\em Proceedings of the IEEE/CVF Conference on Computer Vision and
  Pattern Recognition}, pages 4267--4276, 2021.

\bibitem[\protect\citeauthoryear{Jens \bgroup \em et al.\egroup
  }{2019}]{behley2019iccv}
Behley Jens, Garbade Martin, Milioto Andres, Quenzel Jan, Behnke Sven,
  Stachniss Cyrill, and Gall Jurgen.
\newblock Semantickitti: A dataset for semantic scene understanding of lidar
  sequences.
\newblock In {\em Proceedings of the IEEE/CVF International Conference on
  Computer Vision}, 2019.

\bibitem[\protect\citeauthoryear{Johnson and Hebert}{1999}]{johnson1999using}
Andrew~E Johnson and Martial Hebert.
\newblock Using spin images for efficient object recognition in cluttered 3d
  scenes.
\newblock {\em IEEE Transactions on Pattern Analysis and Machine Intelligence},
  21(5):433--449, 1999.

\bibitem[\protect\citeauthoryear{Li \bgroup \em et al.\egroup
  }{2020}]{li2020iterative}
Jiahao Li, Changhao Zhang, Ziyao Xu, Hangning Zhou, and Chi Zhang.
\newblock Iterative distance-aware similarity matrix convolution with
  mutual-supervised point elimination for efficient point cloud registration.
\newblock In {\em Proceedings of the European Conference on Computer Vision},
  pages 378--394. Springer, 2020.

\bibitem[\protect\citeauthoryear{Lu \bgroup \em et al.\egroup
  }{2021}]{lu2021hregnet}
Fan Lu, Guang Chen, Yinlong Liu, Lijun Zhang, Sanqing Qu, Shu Liu, and Rongqi
  Gu.
\newblock Hregnet: A hierarchical network for large-scale outdoor lidar point
  cloud registration.
\newblock In {\em Proceedings of the IEEE/CVF International Conference on
  Computer Vision}, pages 16014--16023, 2021.

\bibitem[\protect\citeauthoryear{Lucas and Kanade}{1981}]{lucas1981iterative}
Bruce~D Lucas and Takeo Kanade.
\newblock An iterative image registration technique with an application to
  stereo vision.
\newblock In {\em Proceedings of DARPA Image Understanding Workshop}, pages
  121--130. Vancouver, British Columbia, 1981.

\bibitem[\protect\citeauthoryear{Pais \bgroup \em et al.\egroup
  }{2020}]{pais20203dregnet}
G~Dias Pais, Srikumar Ramalingam, Venu~Madhav Govindu, Jacinto~C Nascimento,
  Rama Chellappa, and Pedro Miraldo.
\newblock 3dregnet: A deep neural network for 3d point registration.
\newblock In {\em Proceedings of the IEEE/CVF Conference on Computer Vision and
  Pattern Recognition}, pages 7193--7203, 2020.

\bibitem[\protect\citeauthoryear{Perfecto \bgroup \em et al.\egroup
  }{2017}]{perfecto2017milimeter}
Cristina Perfecto, Javier Del~Ser, and Mehdi Bennis.
\newblock Millimeter-wave v2v communications: Distributed association and beam
  alignment.
\newblock {\em IEEE J-SAC}, 35(9):2148--2162, 2017.

\bibitem[\protect\citeauthoryear{Poiesi and
  Boscaini}{2021}]{poiesi2021distinctive}
Fabio Poiesi and Davide Boscaini.
\newblock Distinctive 3d local deep descriptors.
\newblock In {\em Proceedings of the International Conference on Pattern
  Recognition}, pages 5720--5727. IEEE, 2021.

\bibitem[\protect\citeauthoryear{Qi \bgroup \em et al.\egroup
  }{2017}]{qi2017pointnet}
Charles~R Qi, Hao Su, Kaichun Mo, and Leonidas~J Guibas.
\newblock Pointnet: Deep learning on point sets for 3d classification and
  segmentation.
\newblock In {\em Proceedings of the IEEE/CVF Conference on Computer Vision and
  Pattern Recognition}, pages 652--660, 2017.

\bibitem[\protect\citeauthoryear{Qin \bgroup \em et al.\egroup
  }{2022}]{qin2022geometric}
Zheng Qin, Hao Yu, Changjian Wang, Yulan Guo, Yuxing Peng, and Kai Xu.
\newblock Geometric transformer for fast and robust point cloud registration.
\newblock In {\em Proceedings of the IEEE/CVF Conference on Computer Vision and
  Pattern Recognition}, pages 11143--11152, 2022.

\bibitem[\protect\citeauthoryear{Rusu \bgroup \em et al.\egroup
  }{2009}]{rusu2009fast}
Radu~Bogdan Rusu, Nico Blodow, and Michael Beetz.
\newblock Fast point feature histograms (fpfh) for 3d registration.
\newblock In {\em Proceedings of the IEEE International Conference on Robotics
  and Automation}, pages 3212--3217. IEEE, 2009.

\bibitem[\protect\citeauthoryear{Sarode \bgroup \em et al.\egroup
  }{2019}]{sarode2019pcrnet}
Vinit Sarode, Xueqian Li, Hunter Goforth, Yasuhiro Aoki, Rangaprasad~Arun
  Srivatsan, Simon Lucey, and Howie Choset.
\newblock Pcrnet: Point cloud registration network using pointnet encoding.
\newblock {\em arXiv preprint arXiv:1908.07906}, 2019.

\bibitem[\protect\citeauthoryear{Sun \bgroup \em et al.\egroup
  }{2020}]{sun2020circle}
Yifan Sun, Changmao Cheng, Yuhan Zhang, Chi Zhang, Liang Zheng, Zhongdao Wang,
  and Yichen Wei.
\newblock Circle loss: A unified perspective of pair similarity optimization.
\newblock In {\em Proceedings of the IEEE/CVF Conference on Computer Vision and
  Pattern Recognition}, pages 6398--6407, 2020.

\bibitem[\protect\citeauthoryear{Thomas \bgroup \em et al.\egroup
  }{2019}]{thomas2019kpconv}
Hugues Thomas, Charles~R Qi, Jean-Emmanuel Deschaud, Beatriz Marcotegui,
  Fran{\c{c}}ois Goulette, and Leonidas~J Guibas.
\newblock Kpconv: Flexible and deformable convolution for point clouds.
\newblock In {\em Proceedings of the IEEE/CVF International Conference on
  Computer Vision}, pages 6411--6420, 2019.

\bibitem[\protect\citeauthoryear{Tombari \bgroup \em et al.\egroup
  }{2010}]{tombari2010unique}
Federico Tombari, Samuele Salti, and Luigi Di~Stefano.
\newblock Unique signatures of histograms for local surface description.
\newblock In {\em Proceedings of the European Conference on Computer Vision},
  pages 356--369. Springer, 2010.

\bibitem[\protect\citeauthoryear{Wang and Solomon}{2019a}]{wang2019deep}
Yue Wang and Justin~M Solomon.
\newblock Deep closest point: Learning representations for point cloud
  registration.
\newblock In {\em Proceedings of the IEEE/CVF International Conference on
  Computer Vision}, pages 3523--3532, 2019.

\bibitem[\protect\citeauthoryear{Wang and Solomon}{2019b}]{wang2019prnet}
Yue Wang and Justin~M Solomon.
\newblock Prnet: Self-supervised learning for partial-to-partial registration.
\newblock {\em arXiv preprint arXiv:1910.12240}, 2019.

\bibitem[\protect\citeauthoryear{Wang \bgroup \em et al.\egroup
  }{2020}]{wang2020demystifying}
Song Wang, Jingqi Huang, and Xinyu Zhang.
\newblock Demystifying millimeter-wave v2x: Towards robust and efficient
  directional connectivity under high mobility.
\newblock In {\em Proceedings of the ACM MobiCom}, 2020.

\bibitem[\protect\citeauthoryear{Weixin \bgroup \em et al.\egroup
  }{2019}]{lu1905deepicp}
Lu~Weixin, Wan Guowei, Zhou Yao, Fu~Xiangyu, Yuan Pengfei, and Song Shiyu.
\newblock Deepicp: An end-to-end deep neural network for 3d point cloud
  registration.
\newblock {\em arXiv preprint arXiv:1905.04153}, 2019.

\bibitem[\protect\citeauthoryear{Yew and Lee}{2018}]{yew20183dfeat}
Zi~Jian Yew and Gim~Hee Lee.
\newblock 3dfeat-net: Weakly supervised local 3d features for point cloud
  registration.
\newblock In {\em Proceedings of the European Conference on Computer Vision},
  pages 607--623, 2018.

\bibitem[\protect\citeauthoryear{Yew and Lee}{2020}]{yew2020rpm}
Zi~Jian Yew and Gim~Hee Lee.
\newblock Rpm-net: Robust point matching using learned features.
\newblock In {\em Proceedings of the IEEE/CVF Conference on Computer Vision and
  Pattern Recognition}, pages 11824--11833, 2020.

\bibitem[\protect\citeauthoryear{Yu \bgroup \em et al.\egroup
  }{2021}]{yu2021cofinet}
Hao Yu, Fu~Li, Mahdi Saleh, Benjamin Busam, and Slobodan Ilic.
\newblock Cofinet: Reliable coarse-to-fine correspondences for robust
  pointcloud registration.
\newblock {\em Advances in Neural Information Processing Systems},
  34:23872--23884, 2021.

\bibitem[\protect\citeauthoryear{Zeng \bgroup \em et al.\egroup
  }{2017}]{zeng20173dmatch}
Andy Zeng, Shuran Song, Matthias Nie{\ss}ner, Matthew Fisher, Jianxiong Xiao,
  and Thomas Funkhouser.
\newblock 3dmatch: Learning local geometric descriptors from rgb-d
  reconstructions.
\newblock In {\em Proceedings of the IEEE/CVF Conference on Computer Vision and
  Pattern Recognition}, pages 1802--1811, 2017.

\end{thebibliography}

\end{document}